%% file: image_set_localization.tex
\documentclass[10pt,conference,a4paper]{IEEEtran}
\usepackage{times,amsmath,epsfig}
\usepackage{epstopdf}

\input{common.tex}

\title{Image Set Querying Based Localization}
\author{%
{Lei Deng*, \quad Siyuan Huang*, \quad Yueqi Duan, \quad Baohua Chen, \quad Jie Zhou}\\
\vspace{1.6mm}\\
\fontsize{10}{10}\selectfont\itshape

{Tsinghua National Laboratory for Information Science and Technology} \\
{Department of Automation, Tsinghua University, Beijing, China} \\
\fontsize{9}{9}\selectfont\ttfamily\upshape
}

\begin{document}
\maketitle

\renewcommand{\thefootnote}{}
\footnote{%
\hrule
\vspace{4pt}
* indicates equal contributions
}

\begin{abstract}
Conventional single image based localization methods usually fail to localize a querying image when there exist large variations between the querying image and the pre-built scene. To address this, we propose an image-set querying based localization approach. When the localization by a single image fails to work, the system will ask the user to capture more auxiliary images. First, a local 3D model is established for the querying image set. Then, the pose of the querying image set is estimated by solving a nonlinear optimization problem, which aims to match the local 3D model against the pre-built scene. Experiments have shown the effectiveness and feasibility of the proposed approach.
\end{abstract}


\section{Introduction}

Image-based localization has been widely used in many vision applications such as auto navigation~\cite{pami/DavisonRMS07}, augmented reality~\cite{iswc/CastleKM08}, and photo collection visualization~\cite{tog/SnavelySS06}. The aim of image-based localization is to estimate the camera's pose~(orientation and position) in an interested area from a single querying image. Generally, there are three key steps in a single image-based localization system~\cite{iccv/SattlerLK11,eccv/LiSHF12}: 1) 2D local features~(\emph{e.g.,} SIFT \cite{ijcv/Lowe04}) are extracted from the querying image, 2) matching between local features from the querying image and the 3D point cloud of the scene which also contains corresponding feature descriptors, and 3) camera pose estimation by solving a perspective-n-point (PNP) problem~\cite{cacm/FischlerB81,cvpr/BujnaKKP08,conf/iccv/KukelovaBP13}.

It is challenging to directly match a querying image to the 3D point cloud of the scene, especially when there exist large variations between them. The reason is that the model of the scene is usually built up under a fixed environment which is different to that of the querying image. For instance, a scene is reconstructed by high quality street views, and the surveillance cameras to be localized are working under different illumination conditions and are distant from that of the street view, as shown in \refFig{desc}. Therefore, conventional single image localization methods~\cite{iccv/SattlerLK11,eccv/LiSHF12} fail due to they rely heavily on 2D-3D matches between features, which are hardly to be available in these challenging scenarios.

\begin{figure}[tb]
\centering
\includegraphics[width=0.7\linewidth]{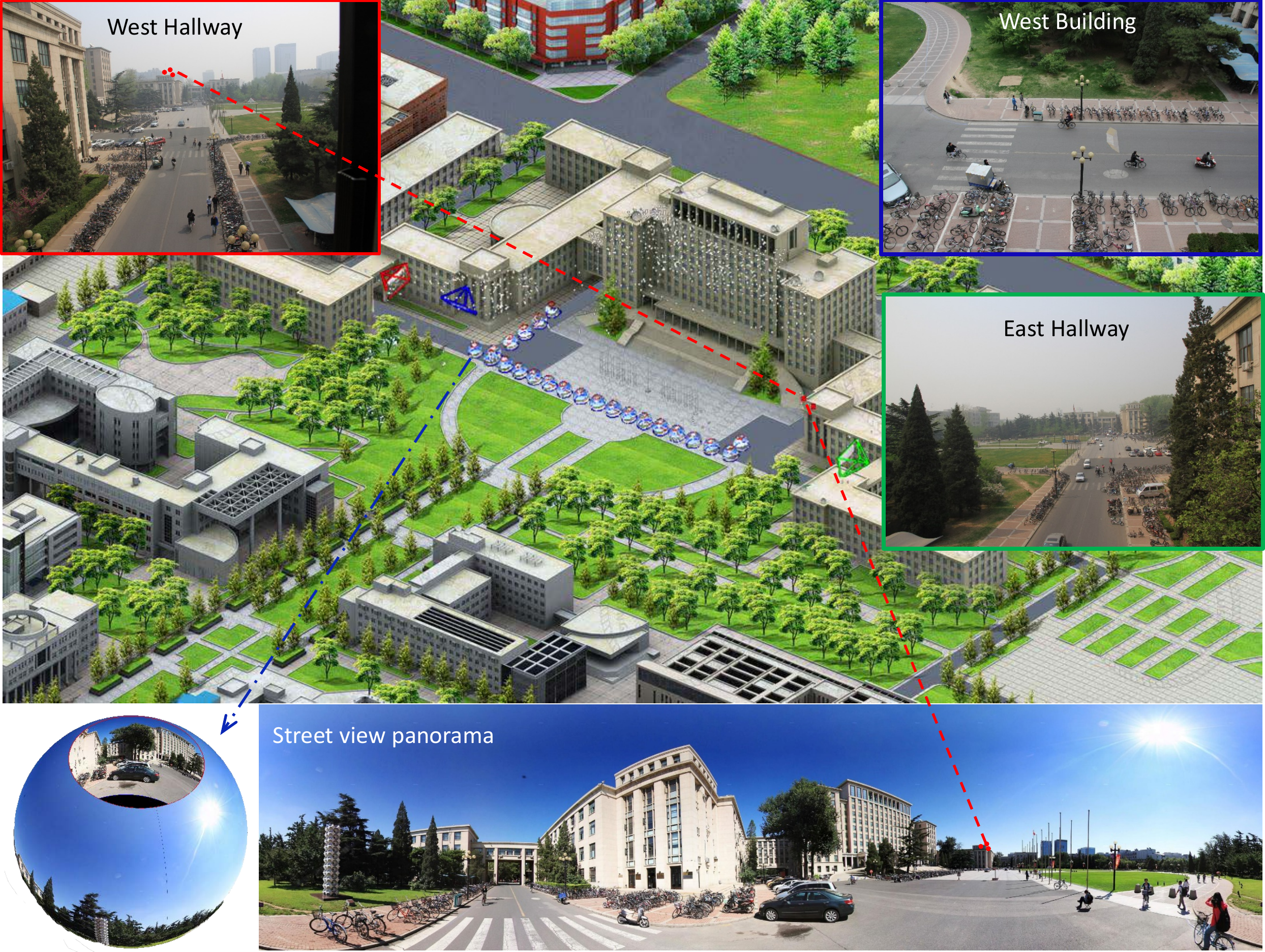}
\caption{\small{Illustration of the challenge of the feature matching. The scene 3D point cloud (yellow dots on the main building) is reconstructed by street view panoramas (bottom). Surveillance cameras from West Hallway (red), West Building (blue) and East Hallway (green) are required to localize. There are only a few feature matches between the West Hallway image and the scene. No feature matches can be found in the West Building image and the East Hallway image due to the large pose variations from the scene.}}
\labelFig{desc}
\end{figure}

To address this, in this paper, we present an image-set querying based localization approach. The framework is shown in \refFig{framework}. When the pose estimation by the conventional single image localization method is unsuccessful, the user are asked to capture more auxiliary images to assist the localization task. Together with the querying image, these bridging images are aggregated to form a local 3D model. Then a 3D-to-3D feature matching scheme is taken to obtain reliable matches between the querying image set and the scene 3D point cloud. The pose of the image set is estimated by solving a nonlinear optimization problem. Besides using the reconstructed camera poses in the local camera set coordinate system, local 3D point information is explored in the nonlinear optimization stage for a further re-projection error minimization. Since the image set not only contains more information for localization, but also has stronger inherent geometry constraints, better localization performance can be obtained.

This paper is organized as follows. \refSec{camera pose} introduces a new camera set pose solver. \refSec{localization} details the proposed image set querying based localization approach. \refSec{experiments} presents the experimental results, and \refSec{conclusion} concludes this paper.

\begin{figure}[tb]
\centering
\includegraphics[width=0.8\linewidth]{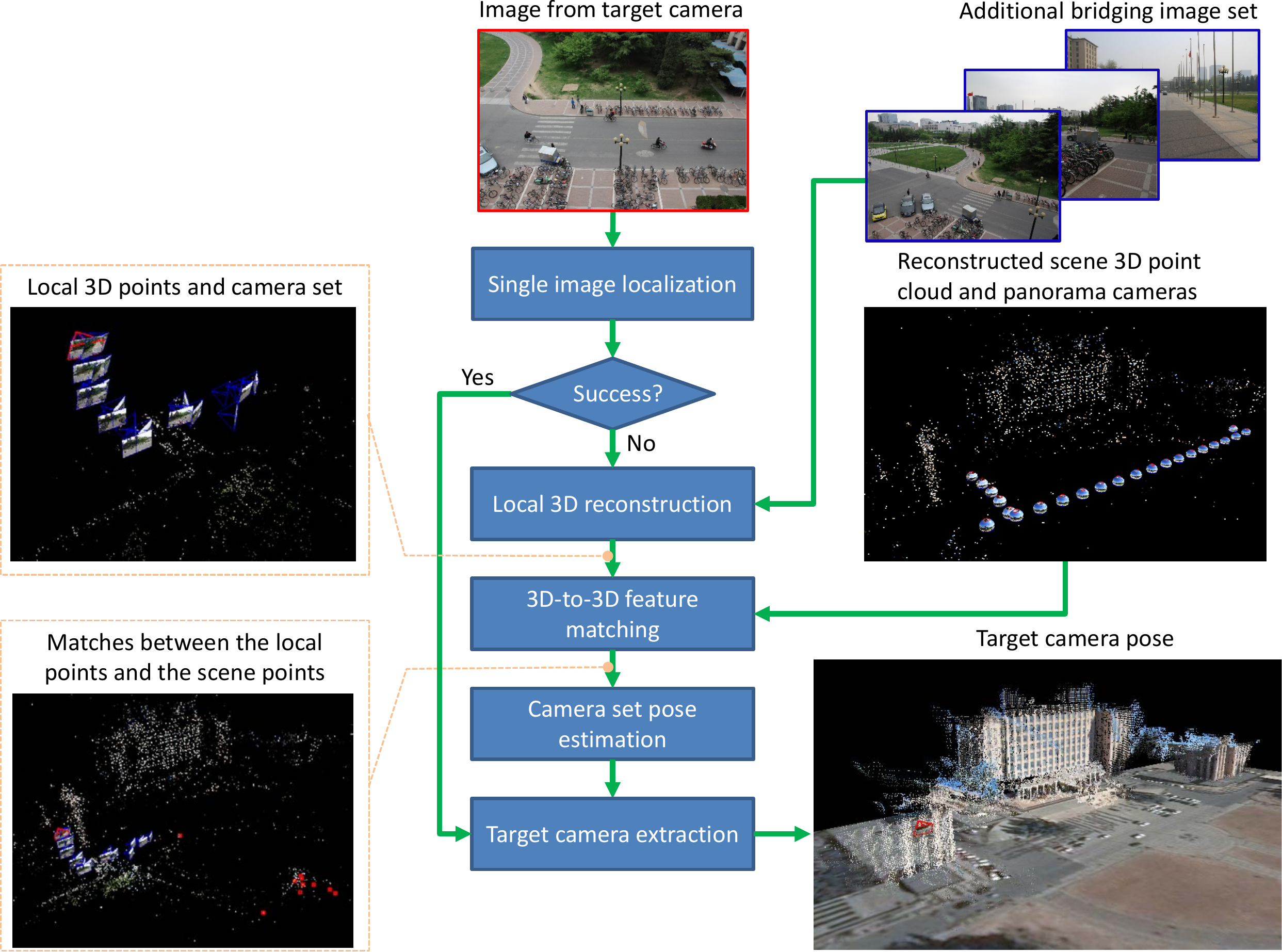}
\caption{\small{The framework of the proposed image set querying based localization.}}
\labelFig{framework}
\end{figure}

% At the beginning, the scene 3D points with their corresponding feature descriptors is built up by a pinhole camera model structure-from-motion procedure. These features are indexed by a kd-tree preparing for the online feature matching stage. When the target image comes and single image localization fails, additional bridging images are needed which forms an image set. Then they are aggregated to form a local 3D model by a local reconstruction procedure. After that, a 3D-to-3D feature matching stage is applied between the image set 3D model and the kd-tree of the pre-built 3D points. Finally, a new generic camera pose solver is used to solve the image set 3D model's pose as a whole following a nonlinear optimization. The target camera's pose can be extracted from the estimated pose of image set 3D model.

\section{Camera set pose estimation}
\labelSec{camera pose}

A set of pinhole cameras could be considered as a generic camera which is represented by a bag of rays. \refFig{camset_pose_estimation} illustrates the basic idea of using a set of pinhole cameras to form a generic camera, where these rays may not come from the same single optical center.

\begin{figure}[tb]
\centering
\includegraphics[width=0.7\linewidth]{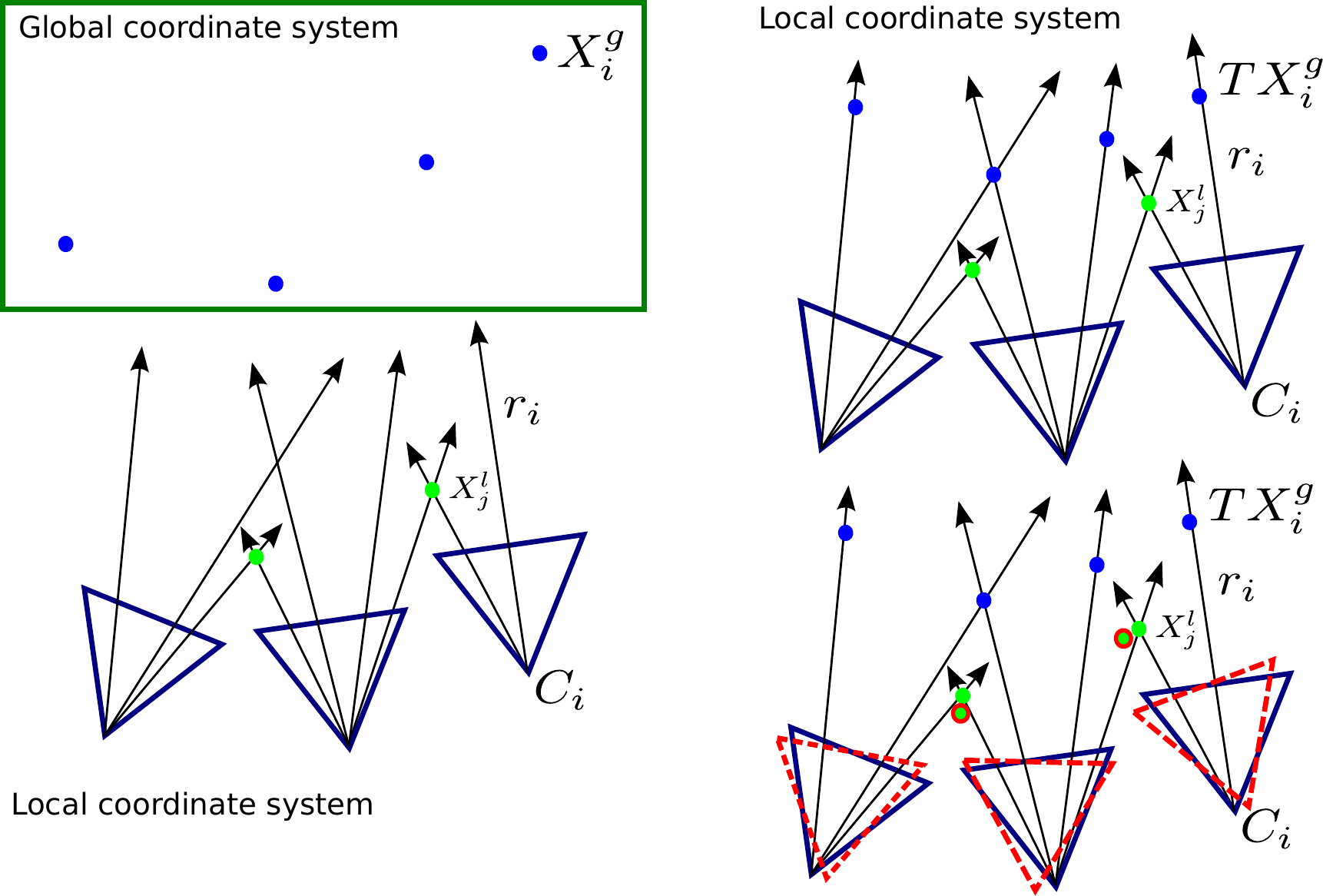}
\caption{\small{A pinhole camera set forms a generic camera and an observed ray $r_i$ of the generic camera comes from center $C_i$ directing at the global fixed 3D point $X^g_i$. By applying a rigid transformation $T$, corresponding 3D point $X^g_i$ in global coordinate system is mapped to local camera set coordinate system as $TX^g_i$ (top right). The rigid transformation $T$ can be seen as the pose of the camera set and can be estimated using the proposed DLT solver. With a nonlinear optimization, the constraints introduced by the 3D point $X^l_j$ in local camera set coordinate system are explored. Relative poses among these cameras $C_i$ and the local 3D points $X^l_j$ could be adjusted (red) for a further re-projection error minimization (bottom right).}}
\labelFig{camset_pose_estimation}
\end{figure}

Similar to the single camera pose estimation, the camera set pose estimation problem can be defined as, given some rays~(direction $r_i$ with its projection center $C_i$) and their corresponding global fixed 3D points $X^g_i$, find the camera set pose which is the rigid transformation~($T=sR[I|-\!\!C]$) to map the matched 3D points from the global coordinate system to the camera set's local coordinate system.

According to the geometry constraints, we have
\begin{equation}
\labelEqn{generic_pnp}
r_i \times (TX^g_i-C_i)=0,
\end{equation}
where $C_i$ is the projection center and $r_i$ is the ray's direction, respectively.

There are 12 unknown variables and 7 DOFs~(6 for pose and 1 for scale) in the transformation~$T$. Direct linear transformation~(DLT) method can be applied to solve the estimation problem. By rearranging \refEqn{generic_pnp}, we have the following equation:
\begin{equation}
\labelEqn{generic_pnp_dlt}
[r_i]_{\times}TX^g_i=[r_i]_{\times}C_i.
\end{equation}
By applying the Kronecker product property, \refEqn{generic_pnp_dlt} can be re-written as
\begin{equation}
{X^g_i}^T\otimes[r_i]_{\times}vec(T)=[r_i]_{\times}C_i.
\end{equation}
Since two independent constraints can be provided by one observation~($r_i, C_i,X^g_i$), at least 6 points are required to solve the problem with DLT. Having obtained the transformation $T$, we need to project the 12-DOF space into a 7-DOF valid similarity transformation space. The $K[R|t]$ from camera matrix $P$ could be decomposed~\cite{hartley06multiple} by the transformation $T$ as
\begin{equation}
K,[R|t]=rq\_decomposition(T),
\end{equation}
Then the valid 7 DOF transform $T_{DLT}$ is projected as
\begin{equation}
T_{DLT}=s[R|t]=\frac{trace(K)}{3}[R|t],
\end{equation}
where the scale factor $s$ is the average value of $K$'s diagonal elements. After that, the Levenberg-Marquardt algorithm is used to minimize the re-projection error, which is the golden standard in geometry estimation \cite{hartley06multiple}. Initializing $T$ as $T_{DLT}$, the optimization objective function is formulated as
\begin{equation}
T_{LM} = \arg \underset{T}{\min}\sum_{i,j}\|r_{ij} - \frac{P_i\widetilde{(T{X^g_j})}}{\|P_i\widetilde{(T{X^g_j})}\|}\|.
\end{equation}

Previous solver and optimization considered the camera set as a whole rigid object, and the relative poses among pinhole cameras could not be changed. However, if the poses of these cameras are also reconstructed by a 3D reconstruction algorithm, relative poses among them may not be accurate. So a further optimization is needed to adjust the inner relative poses for better re-projection error minimization.

Beside the corresponding global 3D points $X^g_i$, local 3D points $X^l_k$ reconstructed in the camera set coordinate system from the image set are also involved in the optimization step. Then the nonlinear optimization becomes the following objective function,

\begin{equation}
\begin{aligned}
T_{OPT} = \arg \underset{T}{\min}&\sum_{i,j}\|r_{ij} - \frac{P_i\widetilde{(T{X^g_j}}}{\|P_i\widetilde{(TX^g_j)}\|}\| +\\
&\sum_{i,j}\|r^l_{ij}-\frac{P_i\widetilde{X^l_j}}{\|P_i\widetilde{X^l_j}\|}\|,
\end{aligned}
\end{equation}
where $r^l_{ij}$ is the observed ray from the $j$-th camera directing at locally reconstructed $j$-th 3D point $X^l_j$. After this optimization, more geometry constraints introduced by $X^l$ can be explored and a better pose estimation can be achieved, as illustrated in \refFig{camset_pose_estimation}.

\section{Image set localization}
\labelSec{localization}
Compared to the ground area, surrounding buildings usually have rich information for localization. For example, if a person doesn't known where he is, he will look around to get his position. To capture such context information of the scene for better localization, 360 panorama is a good choice due to its rich and compact information of surroundings. With the help of existing street view panoramas with high quality at regular distributed locations, high quality scene 3D model is guaranteed to be reconstructed and human efforts can be greatly alleviated. Furthermore, large scale 3D modeling is possible.

Conventional structure-from-motion methods \cite{tog/SnavelySS06,iccv/AgarwalSSSS09} are under the rectilinear camera model assumption by minimizing pixel re-projection errors. To get a unify representation of both panorama and rectilinear cameras, we use the pinhole camera model instead. The pinhole camera model considers each 2D pixel as a ray passing through a single projection center (optical center) which can be represented as a 3D coordinate $\bx(u,v,w)$ lies on the unit spherical surface in the camera coordinate system. Calibration function $\bx=\kappa(\bu,K), \bu=\kappa^{-1}(\bx, K)$ defines the mapping between a ray $\bx(u,v,w)$ and its corresponding pixel $\bu(u,v)$. Eqn. \ref{eqn:panorama_calib},\ref{eqn:rectilinear_calib} are the calibration functions for the panoramic, fisheye and rectilinear cameras respectively.
\begin{eqnarray}
\labelEqn{panorama_calib}
p=\frac{u-u_c}{f}, t=\frac{v-v_c}{f}, \bu_c=(u_c,v_c), \nonumber \\
\kappa(\bu, (f,\bu_c))=(\cos(t)\sin(p),\sin(t),\cos(t)\cos(p)),
\end{eqnarray}
\begin{eqnarray}
\labelEqn{rectilinear_calib}
p=\arctan(\frac{u-u_c}{f}), t=\arctan(\frac{v-v_c}{f}), \nonumber \\
\kappa(\bu, (f,\bu_c))=(\cos(t)\sin(p),\sin(t),\cos(t)\cos(p)),
\end{eqnarray}
where $\bu_c$ is the principle point's pixel coordinate~(for panorama, any point can be principle point theoretically), $f$ is the focal length, $p$ is panning angle around $y$ axis, and $t$ is tilting angle around $x$ axis. The geometry re-projection ray error becomes
\begin{equation}
 \|\kappa(\bu_{ij},K_i)- \frac{P_i\widetilde{X_j}}{\|P_i\widetilde{X_j}\|}\|,
\end{equation}
where $\widetilde{X_j}=(X;1)$ is the homogeneous coordinate of $j$-th 3D point, $P_i=[R,t]$ is the $i$-th camera projection matrix, elements in $K_i$ represent the intrinsic parameters of the $i$-th camera.

According to the geometry properties of the pinhole camera model, conventional rectilinear camera model structure-from-motion building blocks such as two view geometry, triangulation, perspective-n-point and bundle adjustment, should be adjusted. By applying the pinhole model based structure-from-motion procedure, 3D scene point cloud and cameras can be reconstructed, as shown in \refFig{framework}. Each 3D point corresponds to several 2D features~(SIFT), and these 2D features are indexed by a kd-tree method for accelerating the online feature matching stage.

When the querying image comes, we first apply the conventional single image based localization technique. If the pose estimation fails, it means that the querying image has large variations compared to the pre-built scene. Under this case, additional images are required to help matching between the querying image and the scene 3D point cloud. These images can be captured in the area from the target camera to the scene as a bridge. Together with the querying image, these bridging images are aggregated to form a local 3D model (3D point cloud and cameras) by the previous structure-from-motion procedure where the inherent geometry constraints are enhanced. Based on a 3D-to-3D feature matching stage, the image set 3D model is matched against to the scene 3D point cloud. Finally, the pose of querying image set model is estimated by using the camera set pose solver described in \refSec{camera pose} and the target camera's pose can be extracted as shown in \refFig{framework}.

The 3D-to-3D matching stage works as follows: two nearest neighbors in the 3D point cloud of the scene for each local 3D point in the image set are first identified. Then, the ratio of the distance between the local 3D point and the nearest neighbor and the second nearest neighbor are tested. At last, the ratio test is employed reversely to filter out bad local 3D points to get enough reliable 3D-to-3D feature matches.

\begin{table*}[tb]
\vspace{5pt}
\centering
\captionsetup{belowskip=-0pt,aboveskip=0pt}
\caption{\small{Successful registration rate in the three querying image sets.}}
\begin{tabular}{c|cc|cc|cc}
\hline
method & \multicolumn{2}{c|}{West Hallway} & \multicolumn{2}{c|}{East Hallway} & \multicolumn{2}{c}{West Building}\\
& \#reg./\#total & \#reg. rate & \#reg./\#total & \#reg. rate & \#reg./\#total & \#reg. rate\\
\hline
single image based\cite{iccv/SattlerLK11} & 4/14 & 28.57\% & 12/15 & 80.00\% & 1/21 & 4.76\%\\
\hline
proposed & 14/14 & 100.00\% & 15/15 & 100.00\% & 21/21 & 100.00\%\\
\hline
\end{tabular}
\labelTab{reg_rate}
\vspace{9pt}
\end{table*}

\begin{table*}[tb]
\centering
\captionsetup{belowskip=-0pt,aboveskip=0pt}
\caption{\small{Evaluation of the location error on image set and target image. The statistical results for single querying image based method \cite{iccv/SattlerLK11} is from successful registered images only.}}
\begin{tabular}{c|c|ccccc|c}
\hline
\multicolumn{2}{c|}{} & \multicolumn{5}{c|}{image set} & target image \\
\hline
\multirow{2}{*}{dataset} & \multirow{2}{*}{method} & \#reg./\#total & min  & median & max  & mean & recon. err \\
&&& (m/deg) & (m/deg) & (m/deg) & (m/deg) & (m/deg)\\
\hline
\multirow{3}{*}{West Hallway} & single image based\cite{iccv/SattlerLK11} & 4/14 &2.081/1.001 & 2.708/1.019 & 5.029/1.092 & 3.131/1.033 & 5.028/1.092\\
&camset & 14/14 & 2.175/0.949 & 3.031/0.999 & 3.973/1.546 & 3.030/1.042 & 3.973/1.092\\
&camset+opt & 14/14&1.770/0.915 &2.455/0.970 & 3.248/1.546 &\textbf{2.457/1.042}&\textbf{3.248/1.095} \\
\hline
\multirow{3}{*}{East Hallway} &single image based\cite{iccv/SattlerLK11}& 12/15 & 1.015/0.335 & 3.672/1.278 & 23.014/7.815 & 8.418/2.827 & -\\
&camset & 15/15 & 2.833/1.116 & 3.297/1.129 & 3.697/1.154 & 3.293/1.1.131 & 3.641/1.154\\
&camset+opt & 15/15 & 2.489/0.980 & 2.908/0.991 & 3.273/1.104 & \textbf{2.904/0.991} & \textbf{3.273/1.001} \\
\hline
\multirow{3}{*}{West Building} &single image based\cite{iccv/SattlerLK11}& 1/21 & 4.729/3.847 & 4.729/3.847 &4.729/3.847 &4.729/3.847 &-\\
&camset & 21/21 & 2.167/2.402 & 4.516/3.334 & 4.889/3.658 & 4.247/3.320 & 4.664/3.414\\
&camset+opt & 21/21 & 1.638/2.418 &4.463/2.925 &4.963/3.347 &\textbf{4.197/2.960} &\textbf{4.487/2.979} \\
\hline
\end{tabular}
\labelTab{loc results}
\end{table*}

\section{Experiments}
\labelSec{experiments}

We build scene 3D point cloud for the Main Building in Tsinghua university with hundreds of meters size, which consists of 23 street view panoramas, 3067 3D points and 14330 feature descriptors. The 3D-to-3D ratio test threshold is set to 0.6 and the scene 3D point cloud kd-tree is built by FLANN \cite{flann_pami_2014} with 95\% accuracy. Three image sets are tested, West Hallway (14 images), West Building (15 images) and East Building (21 images).  The querying image is distant to the scene. Experiments are conducted on methods including conventional single image based method \cite{iccv/SattlerLK11}, the proposed camera set pose estimation with and without nonlinear optimization (camset, camset+opt). To evaluate the accuracy of the localization, as done in ~\cite{iccv/SattlerLK11}, a whole 3D model reconstructed by using all images is taken as the ground truth and all the localization results are further checked manually. The minimal 2D/3D matched inliers for conventional single image based localization is 12 (same as \cite{iccv/SattlerLK11}) below which pose estimation is regarded as failed.

\begin{figure}[tb]
\centering
\includegraphics[width=0.8\linewidth]{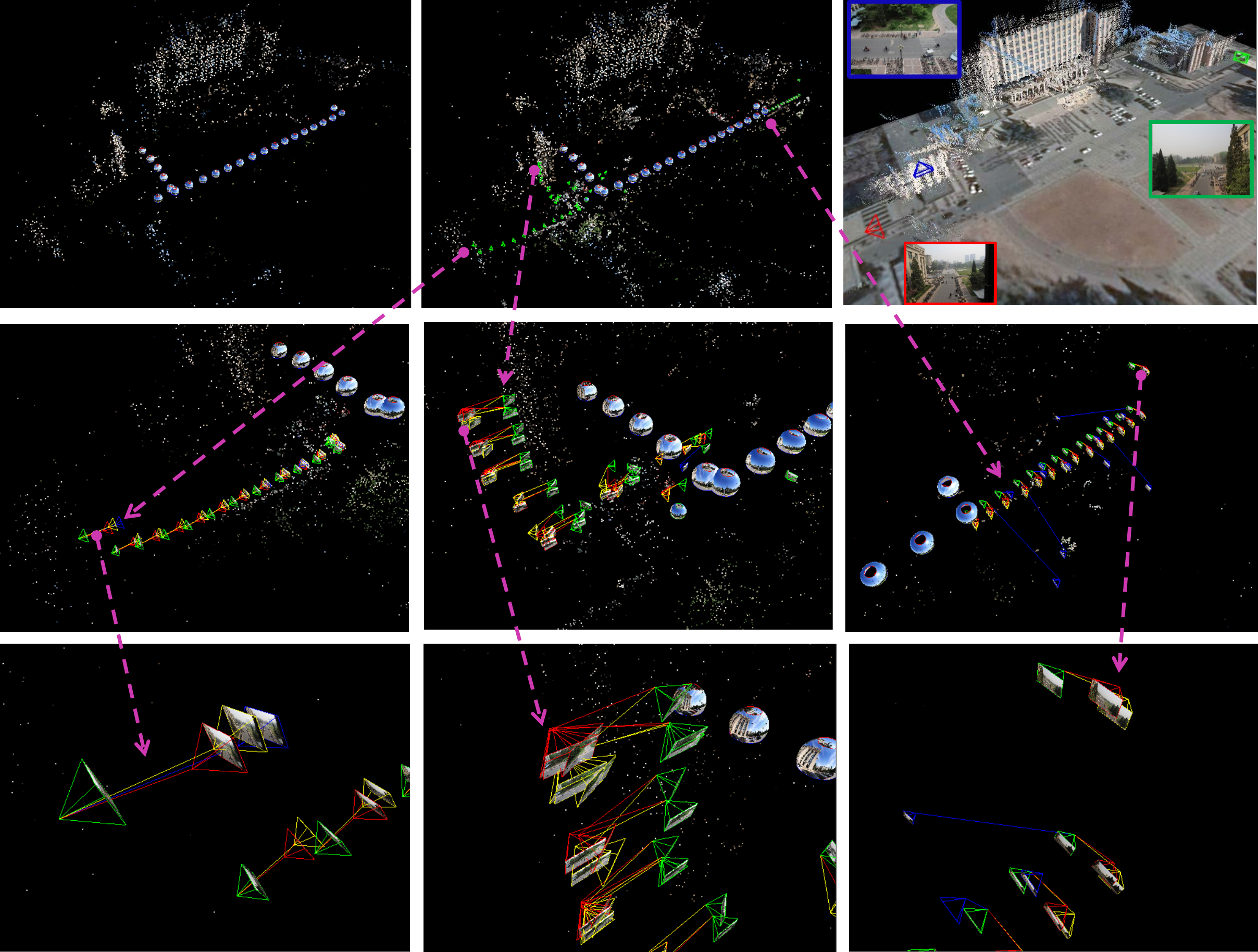}
\caption{\small{Illustration of localization results. Top left shows the reconstructed scene in main building dataset. Top middle is the reconstructed ground truth scene using all three image sets. Top right is the final localization result. Bottom two rows show details of the West Hallway, West Building and East Hallway image sets with different methods, single image based\cite{iccv/SattlerLK11} (blue), camset (yellow) and camset+opt (red). Images in the bottom row are enlarged from the images in the top row. We link the ground truth with corresponding estimated camera center to visualize the displacement.}}
\labelFig{loc results}
\end{figure}

Conventional single image based localization method \cite{iccv/SattlerLK11} cannot estimate the poses of the cameras in most cases due to the environment that they are taken in is different to that of the pre-built scene. The proposed methods can locate each image in the querying image set successfully, as shown in \refTab{reg_rate}.

The qualitative localization results are shown in \refFig{loc results} and the quantitative results are listed in \refTab{loc results}. From which we can see that, 1) the querying image is successfully extracted by the proposed framework, and the orientation errors are usually very small (less than 4\textdegree), 2) only a few images can be localized by the conventional single image localization method, while all of them can be successfully localized by the proposed approach. 3) The location error of the proposed methods is smaller than the conventional single image based method, and the performance can be improved when the nonlinear optimization is further applied.

% The qualitative localization results are shown in \refFig{experiment} and the quantitative results are listed in \refTab{experiment}. We make the following observations from these results:

% \begin{itemize}
% \item The target images with large variations are successfully extracted by our framework, and the orientation errors are usually very small (less than 4\textdegree).

% \item Only a few images can be localized by the conventional single image localization method. However, all of them can be successfully localized by our approach.

% \item The location error of the proposed methods is much smaller than the single image method, and the performance can be improved when the nonlinear optimization technique is applied.

% \end{itemize}

\section{Conclusion}
\labelSec{conclusion}

In this paper, we have proposed a new framework to solve the problem of image-based localization by querying a bridging image set rather than a single image. Compared with the single image, the image set not only contains more information for localization with more feature matches, but also has stronger inherent geometry constraints enforced by a local reconstruction. Therefore, it can be employed for improved localization performance. Experimental results have shown the effectiveness and feasibility of the proposed approach. In the future, we will study the way to capture the bridging image set efficiently to further improve the efficiency of the proposed approach.

\section{Acknowledgement}
\small{This work is supported by the National Natural Science
Foundation of China under Grants 61225008, 61373074 and
61373090, the National Basic Research Program of Chi-
na under Grant 2014CB349304, the Ministry of Education
of China under Grant 20120002110033, and the Tsinghua
University Initiative Scientific Research Program.}

% \newpage\hbox{}
% \simpleFigureTop{local_str_result.pdf}{local structure result}{The localization result of previous built image set 3D model. 3D-to-3D matches are shown in left top. And the rest three show the target camera pose in the form of FOV on the ground. The dense 3D point cloud of main building scene for visualization is reconstructed using PMVS \cite{pami/FurukawaP10}.}

\bibliographystyle{IEEEbib}
\bibliography{refs2}

\end{document}

%% file: common.tex
\usepackage{graphicx}
\usepackage{textcomp}
\usepackage{subfig}
\usepackage{amsmath}
\usepackage{bm}
\usepackage{xargs}
\usepackage{algorithm}
\usepackage{algpseudocode}
\usepackage{multirow}
\usepackage{color}
\graphicspath{{res/}}

\newcommand{\labelFig}[1]{\label{fig:#1}}
\newcommand{\refFig}[1]{Fig.~\ref{fig:#1}}
\newcommand{\labelEqn}[1]{\label{eqn:#1}}
\newcommand{\refEqn}[1]{Eqn.~\ref{eqn:#1}}
\newcommand{\labelSec}[1]{\label{sec:#1}}
\newcommand{\refSec}[1]{Section \ref{sec:#1}}

\newcommand{\labelTab}[1]{\label{tab:#1}}
\newcommand{\refTab}[1]{Tab.~\ref{tab:#1}}
\newcommandx\includeImageLineWidth[2][1=1.0]{\includegraphics[width=#1\linewidth]{#2}}
\newcommandx\simpleFigure[4][1=1.0]{
  \begin{figure}[t]
    \centering
    \includeImageLineWidth[#1]{#2}
    \caption{#4}
    \labelFig{#3}
  \end{figure}
}

\newcommandx\simpleFigureTop[4][1=1.0]{
  \begin{figure}[t!]
    \centering
    \includeImageLineWidth[#1]{#2}
    \caption{#4}
    \labelFig{#3}
  \end{figure}
}

\newcommandx\simpleFigureHere[4][1=1.0]{
  \begin{figure}[!h]
    \centering
    \includeImageLineWidth[#1]{#2}
    \caption{#4}
    \labelFig{#3}
  \end{figure}
}

\newcommandx\simpleFigureW[4][1=1.0]{
  \begin{figure*}[t]
    \centering
    \includeImageLineWidth[#1]{#2}
    \caption{#4}
    \labelFig{#3}
  \end{figure*}
}

\newcommandx\simpleFigureWHere[4][1=1.0]{
  \begin{figure*}[!h]
    \centering
    \includeImageLineWidth[#1]{#2}
    \caption{#4}
    \labelFig{#3}
  \end{figure*}
}

\newcommand{\bx}{\bm{x}}
\newcommand{\bu}{\bm{u}}